\documentclass[a4paper, 10 pt, conference]{ieeeconf}
\IEEEoverridecommandlockouts
\usepackage{cite}
\usepackage{amsmath,amssymb,amsfonts}
\usepackage{algorithmic}
\usepackage{graphicx}
\usepackage{tabularx}
\usepackage{multirow}
\usepackage{tikz}
\usetikzlibrary{3d}
\usetikzlibrary{calc}
\usepackage{textcomp}
\usepackage{xcolor}
\usepackage{algorithm}
\usepackage{algorithmic}
\usepackage{subcaption}
\usepackage{mathtools}
\usepackage{url}
\PassOptionsToPackage{bookmarks=false}{hyperref}
\usepackage{hyperref}
\usepackage{scalerel}

\makeatletter
\newcommand{\thickhline}{%
	\noalign {\ifnum 0=`}\fi \hrule height 1pt
	\futurelet \reserved@a \@xhline
}
\newcolumntype{"}{@{\hskip\tabcolsep\vrule width 1pt\hskip\tabcolsep}}

\makeatother

\title{\LARGE \bf
	Rapidly-Exploring Random Graph Next-Best View Exploration for Ground Vehicles
}

\author{Marco Steinbrink$^{1}$, Philipp Koch$^{2}$, Bernhard Jung$^{1}$ and Stefan May$^{2}$
	\thanks{This activity has received funding from the European Institute of Innovation and Technology (EIT), a body of the European Union, under the Horizon 2020, the EU Framework Programme for Research and Innovation.}
	\thanks{$^{1}$Marco Steinbrink and Bernhard Jung are with the Institute for Informatics, Technical University Bergakademie Freiberg, 09599 Freiberg, Germany
		{\tt\small marco.steinbrink@doktorand.tu-freiberg.de}}%
	\thanks{$^{2}$Philipp Koch and Stefan May are with the Faculty of Electrical Engineering, Precision Engineering, Information Technology, Nuremberg Institute of Technology Georg Simon Ohm, 90489 Nuremberg, Germany
		\newline 978-1-6654-1213-1/21/\$31.00 \textcopyright 2021 IEEE}%
}

\begin{document}

\maketitle
\thispagestyle{empty}
\pagestyle{empty}

\begin{abstract}
In this paper, a novel approach is introduced which utilizes a Rapidly-exploring Random Graph to improve sampling-based autonomous exploration of unknown environments with unmanned ground vehicles compared to the current state of the art. Its intended usage is in rescue scenarios in large indoor and underground environments with limited teleoperation ability. Local and global sampling are used to improve the exploration efficiency for large environments. Nodes are selected as the next exploration goal based on a gain-cost ratio derived from the assumed 3D map coverage at the particular node and the distance to it. The proposed approach features a continuously-built graph with a decoupled calculation of node gains using a computationally efficient ray tracing method. The Next-Best View is evaluated while the robot is pursuing a goal, which eliminates the need to wait for gain calculation after reaching the previous goal and significantly speeds up the exploration. Furthermore, a grid map is used to determine the traversability between the nodes in the graph while also providing a global plan for navigating towards selected goals. Simulations compare the proposed approach to state-of-the-art exploration algorithms and demonstrate its superior performance.
\end{abstract}

\section{Introduction}
The use of robots to undertake a first assessment in dangerous situations like f.e. fires or mine cave-ins has significantly increased during the last decade, a prominent example was the firefighter robot that entered the burning cathedral Notre-Dame in Paris, France in 2019. During these incidents, the robots are generally remote-controlled which limits their use to operations with a sufficient signal transmission. Autonomous agents would facilitate obtaining an overview without putting humans at risk in these situations.

\begin{figure}[htbp]
	\centering
	\begin{subfigure}[t]{.24\textwidth}
		\centering
		\includegraphics[width=\textwidth]{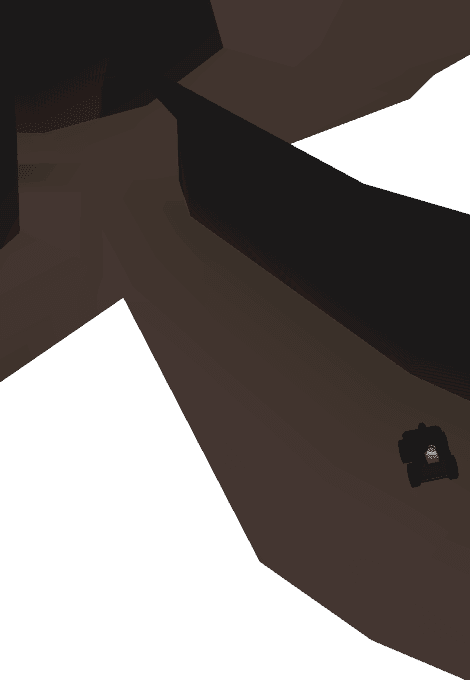}
	\end{subfigure}%
	\hfill
	\begin{subfigure}[t]{.24\textwidth}
		\centering
		\includegraphics[width=\textwidth]{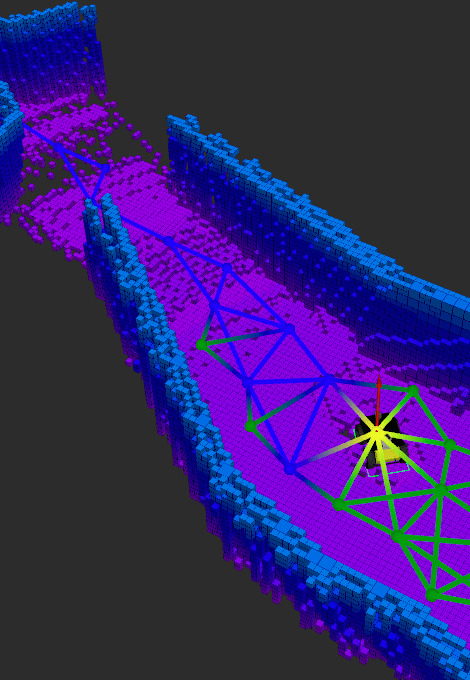}
	\end{subfigure}
	\caption{Simulated underground cave on the left with the RRG and OctoMap on the right. The OctoMap is color encoded by height and truncated at 2m height for visibility. The RRG is colored in blue for nodes worth exploring, green for nodes without sufficient gain and yellow for the current goal.}
	\label{fig:cave}
	\vspace{-5mm}
\end{figure}

Therefore, this paper presents an approach called Rapidly-Exploring Random Graph Next-best View Exploration (RNE) which employs a Rapidly-Exploring Random Graph (RRG) to randomly sample points in the environment to be autonomously explored by an unmanned ground vehicle (UGV). The gain calculation for the sampled points is based on ray tracing in a voxel map to identify the Next-Best View (NBV) which is selected as a goal. 2D traversability grid maps are used for collision avoidance and path planning for the UGV. Simulations compare the proposed approach to state-of-the-art, sampling-based algorithms in different environments, e.g. the exploration of a large underground environment which can be seen in Fig. \ref{fig:cave}.

The following contributions are shown in this work: 
\begin{itemize}
	\item A single, continuously-built RRG to achieve global coverage while providing collision-free, traversable paths.
	\item Decoupled gain calculation which means that each sample's gain is calculated in a separate thread while new goals can be selected without information on all samples. An interrupt mechanism enables to replace the current goal with a better one which gain was just calculated. This aims to decrease exploration time due to removing the common stop-and-compute method of recent state-of-the-art approaches \cite{Bircher2016}\cite{Selin2019}.
	\item RNE is openly available as a Robot Operating System (ROS)\cite{Quigley2009} package for reference and comparison \footnote{\url{https://github.com/MarcoStb1993/rnexploration}}.
\end{itemize}

\section{Related Work}\label{relatedwork}

The proposed approach utilizes RRG, which was introduced for path planning by \cite{Karaman2011a}. RRG is based on the Rapidly-Exploring Random Tree (RRT) \cite{LaValle1998} which it combines with parts of the learning phase of Probabilistic Roadmaps (PRM) \cite{Kavraki1996}. Both methods are also meant for path planning and have remained an active research topic, including many improvements to the computational efficiency and generated path optimality \cite{Yershova2005}\cite{Rickert2008}\cite{Barfoot2014}.

Frontier Exploration (FE) \cite{Yamauchi1997} was one of the earliest  approaches for robotic exploration. More recently, sampling-based methods like RRT gained increased attention in robotic exploration research because they require less computation compared to FE, especially when applied to large 3D problems. This enables their use for online exploration on physically constrained mobile agents. For example, \cite{Umari2017} and \cite{Fang2019} repeatedly construct RRTs to detect frontiers solely in 2D. Our method is aimed at increasing the 3D map coverage for which we also propose a continuously-built graph.

\cite{Vasquez-Gomez2018} applied a process similar to RRT for object reconstruction by evaluating the samples to extract NBVs. RRT was also used to generate efficient inspection paths utilizing ray tracing to estimate the object coverage and calculating a shortest path through the sample points \cite{Bircher2017a} \cite{Song2017}. These approaches are used for known environments while we aim to explore unknown areas which requires approximative gain and cost calculation to define an NBV.

\cite{Bircher2016} introduced the Receding Horizon Next-Best-View Planner (RH-NBVP) which deploys RRT to find NBVs using ray casting. It was designed for autonomous exploration with Unmanned Aerial Vehicles (UAV). RH-NBVP follows RRT's branch with the most potential gain and rebuilds the tree after reaching the first node in the branch. The Autonomous Exploration Planner (AEP) combines RH-NBVP for local exploration with an FE planner for global exploration to avoid premature termination when RH-NBVP finds no branch with a sufficient gain \cite{Selin2019}. Our proposed approach extends and combines AEP's improved gain estimation with a consistent graph that eliminates the rebuilding of the node structure at each iteration to increase the exploration speed as can be seen in the simulations in section \ref{simulations}.

Another approach was proposed by \cite{Schmid2020}, which builds a continuous RRT that is constantly rewired when adding new samples or reaching a goal to ensure, that all nodes are connected to the root through the shortest path. The RRG deployed in our approach makes the rewiring unnecessary as all connections are stored in the graph.

Dang et al. \cite{Dang2019} introduce GBPlanner, an autonomous path planning that builds a local and a global RRG to explore large subterranean environments. The local RRG is constrained by a sliding bounding box and at each iteration the best local path is selected using a combination of path length and expected gain for all nodes on the path. The global planner incorporates all best paths to create a global RRG for homing and intelligent backtracking when the local planner gets stuck.

In \cite{Xu2021}, the Dynamic Exploration Planner (DEP) utilizes a consistent PRM with gain estimation based on RH-NBVP and adds constraint-optimized path planning for the selected trajectory as well as dynamic obstacle avoidance. Compared to GBPlanner and DEP, our proposed approach makes it unnecessary to wait until all gains are computed after reaching the previous goal, as the gain calculation runs in a separate thread.

\section{Problem Description}\label{problem}

Autonomous exploration is intended to map an area $V\in\mathbb{R}^{3}$ which is partitioned into unknown $V_{un}$, free $V_{free}$ and occupied $V_{oc}$ space ($V=V_{un}\cup V_{free}\cup V_{oc}$). An agent aims to classify the initially unknown environment. The explorable space $V_{ex}=V\setminus V_{nx}$ is restricted by the agent's capabilities which determine the non-explorable space $V_{nx}$. During exploration, the already explored space, which is known to the agent, is labeled $V_{av}\in V_{ex}$.

A discrete representation for $V$ is a voxel grid where each voxel has a predefined edge length $e_{V}$ and represents a fraction of $V$. A 2D grid map can be derived from it.

The agent commonly has a limited range which necessitates an efficient strategy to classify $V_{ex}$. Therefore, the NBV for the agent is determined by a function $f(G,C)$. Gain $G$ is derived from the expected additional map coverage at the regarded position and cost $C$ from the agent's distance to the position. The NBV with the best gain-cost ratio (GCR) acquired from $f(G,C)$ will be the agent's next goal. This goal has to be chosen online for autonomous exploration.

\section{Proposed Approach}\label{implementation}

The RRG algorithm, which is based on RRT, had to be adapted for the presented exploration approach. This adaption, its steer function and details regarding the calculation of the GCR are explained in the following.

\subsection{RRG Exploration Algorithm}\label{rrg_exploration}

The algorithm to construct RRG $G=(N,E)$ for exploration is shown in Algorithm \ref{alg:rrt_algorithm}. It requires the robot's position $\vec{x}_{pos}$, a minimum $d_{min}$ and maximum distance $d_{max}$.

\begin{algorithm}[H]
	\caption{RRG expansion}
	\label{alg:rrt_algorithm}
	\begin{algorithmic}[1]
		\renewcommand{\algorithmicrequire}{\textbf{Input:}}
		\REQUIRE $\vec{x}_{pos},d_{min},d_{max}$
		\STATE $G \leftarrow $\texttt{initGraph(}$\vec{x}_{pos}$\texttt{)}
		\WHILE{!\texttt{explorationFinished()}}
		\STATE $V_{av} \leftarrow$ \texttt{retrieveMap()}
		\STATE $\vec{x}_{rand} \leftarrow $\texttt{randomlySamplePoint(}$V_{av}$\texttt{)}
		\STATE $\vec{x}_{n}$ $\leftarrow $ \texttt{findNearestNeighbour(}$G, \vec{x}_{rand}$\texttt{)} 
		\IF{\texttt{d(}$\vec{x}_{rand},\vec{x}_{n}$\texttt{)}$\ge d_{min}$}
		\STATE $\vec{x}_{rand} \leftarrow $ \texttt{alignSamplePoint(}$\vec{x}_{rand},d_{max}$\texttt{)}
		\STATE $N_{d} \leftarrow $ \texttt{findNodesInRadius(}$G, \vec{x}_{rand},d_{max}$\texttt{)}
		\STATE $N_{c} \leftarrow \emptyset$
		\FOR{$\vec{x}_d \in N_{d}$}		
		\STATE $N_{c} \leftarrow N_{c}\: \cup $ \texttt{steer(}$V_{av},\vec{x}_{rand},\vec{x}_d$\texttt{)}	
		\ENDFOR
		\IF{$N_{c} \ne \emptyset$}
		\STATE \texttt{addNodeToGraph(}$G,\vec{x}_{rand},N_{c}$\texttt{)} 
		\ENDIF
		\ENDIF
		\ENDWHILE
	\end{algorithmic}
\end{algorithm}

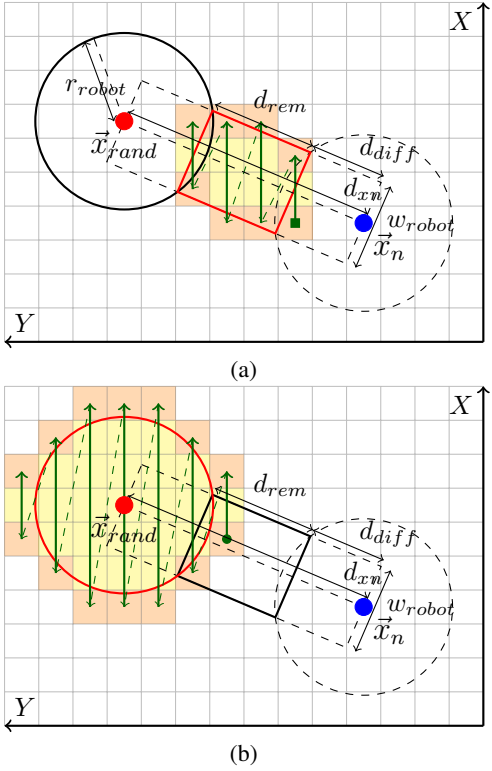
\begin{figure}[htbp]
		\centering
	\begin{subfigure}[t]{.45\textwidth}
		\centering
		\begin{tikzpicture}[scale=0.45]
		\fill [orange,opacity=0.3] (9,3) rectangle (9,4);	
		\fill [orange,opacity=0.3] (6,3) rectangle (9,4);	
		\fill [orange,opacity=0.3] (8,5) rectangle (9,6);	
		\fill [yellow,opacity=0.3] (8,4) rectangle (9,5);	
		\fill [yellow,opacity=0.3] (6,4) rectangle (8,6);
		\fill [orange,opacity=0.3] (5,6) rectangle (8,7);
		\fill [orange,opacity=0.3] (5,4) rectangle (6,5);
		\fill [yellow,opacity=0.3] (5,5) rectangle (6,6);			
		\draw[step=1.0,gray,thin,opacity=0.5] (0,0) grid (14,10);
		\draw[thick,->] (14,0) -- (0,0) node[above right]{$Y$};
		\draw[thick,->] (14,0) -- (14,10) node[below left]{$X$};
		\draw[black, thick] (3.5,6.5) circle (2.6);
		\draw[black,dashed] (10.5,3.5) circle (2.6);
		\draw[black,dashed] ({3.5+1.3*cos(66.8)},{6.5+1.3*sin(66.8)}) -- ({10.5+1.3*cos(66.8)},{3.5+1.3*sin(66.8)}) -- ({10.5-1.3*cos(66.8)},{3.5-1.3*sin(66.8)}) -- ({3.5-1.3*cos(66.8)},{6.5-1.3*sin(66.8)}) -- cycle;
		\draw[red, thick]
		 ({3.5+1.3*cos(66.8)+2.252*cos(23.2)},{6.5+1.3*sin(66.8)-2.252*sin(23.2)}) -- ({10.5+1.3*cos(66.8)-2.252*cos(23.2)},{3.5+1.3*sin(66.8)+2.252*sin(23.2)}) -- ({10.5-1.3*cos(66.8)-2.252*cos(23.2)},{3.5-1.3*sin(66.8)+2.252*sin(23.2)}) -- ({3.5-1.3*cos(66.8)+2.252*cos(23.2)},{6.5-1.3*sin(66.8)-2.252*sin(23.2)}) -- cycle;
		\filldraw[black!60!green,->,thick] (8.5,3.5) -- (8.5,5.5);
		\filldraw[black!60!green,->,dashed] (8.5,5.5) -- (7.5,3.5);
		\filldraw[black!60!green,->,thick] (7.5,3.5) -- (7.5,6.5);
		\filldraw[black!60!green,->,dashed] (7.5,6.5) -- (6.5,3.5);
		\filldraw[black!60!green,->,thick] (6.5,3.5) -- (6.5,6.5);
		\filldraw[black!60!green,->,dashed] (6.5,6.5) -- (5.5,4.5);
		\filldraw[black!60!green,->,thick] (5.5,4.5) -- (5.5,6.5);
		\filldraw[black!60!green] (8.375,3.375) rectangle (8.625,3.625);
		\draw[black,dashed] (3.5,6.5) -- ({3.5+2.6*cos(110)},{6.5+2.6*sin(110)});
		\draw[black,<->] (3.2,6.5) -- ({3.2+2.5*cos(110)},{6.5+2.5*sin(110)});
		\draw[black] (1.5,7) node[anchor=south west] {$r_{robot}$};
		\draw[black,dashed] (3.5,6.5) -- (10.5,3.5);
		\draw[black,<->] ({3.5+0.3*cos(66.8)},{6.5+0.3*sin(66.8)}) -- ({10.5+0.3*cos(66.8)},{3.5+0.3*sin(66.8)});
		\draw[black] (10.5,3.9) node[above] {$d_{xn}$};
		\draw[black,<->] ({10.5+1.3*cos(66.8)+0.3*cos(23.2)},{3.5+1.3*sin(66.8)-0.3*sin(23.2)}) -- ({10.5-1.3*cos(66.8)+0.3*cos(23.2)},{3.5-1.3*sin(66.8)-0.3*sin(23.2)});
		\draw[black] (10.9,4) node[anchor=north west] {$w_{robot}$};
		\draw[black,<->]  ({3.5+1.3*cos(66.8)+2.252*cos(23.2)+0.2*cos(67.8)},{6.5+1.3*sin(66.8)-2.252*sin(23.2)+0.2*sin(67.8)}) -- ({10.5+1.3*cos(66.8)-2.252*cos(23.2)+0.2*cos(67.8)},{3.5+1.3*sin(66.8)+2.252*sin(23.2)+0.2*sin(67.8)});
		\draw[black] (7,6.5) node[anchor=south west] {$d_{rem}$};
		\draw[black,<->] ({10.5+1.3*cos(66.8)-2.252*cos(23.2)+0.2*cos(67.8)},{3.5+1.3*sin(66.8)+2.252*sin(23.2)+0.2*sin(67.8)}) -- ({10.5+1.3*cos(66.8)+0.2*cos(67.8)},{3.5+1.3*sin(66.8)+0.2*sin(67.8)});
		\draw[black] (10.1,5.0) node[anchor=south west] {$d_{diff}$};
		\filldraw[red] (3.5,6.5) circle (0.25) node[below,black]{\large $\vec{x}_{rand}$};
		\filldraw[blue] (10.5,3.5) circle (0.25) node[anchor=north west,black]{\large $\vec{x}_{n}$};
		\end{tikzpicture}
		\caption{}
		\label{fig:collision1}
	\end{subfigure}%
	\hfill
	\begin{subfigure}[t]{.45\textwidth}
		\centering
		\begin{tikzpicture}[scale=0.45]
		\fill [orange,opacity=0.3] (0,5) rectangle (1,6);
		\fill [orange,opacity=0.3] (0,7) rectangle (1,8);
		\fill [orange,opacity=0.3] (1,4) rectangle (2,5);
		\fill [orange,opacity=0.3] (1,8) rectangle (2,9);
		\fill [orange,opacity=0.3] (2,3) rectangle (5,4);
		\fill [orange,opacity=0.3] (2,9) rectangle (5,10);
		\fill [orange,opacity=0.3] (5,4) rectangle (6,5);
		\fill [orange,opacity=0.3] (5,8) rectangle (6,9);
		\fill [orange,opacity=0.3] (6,5) rectangle (7,6);
		\fill [orange,opacity=0.3] (6,7) rectangle (7,8);	
		\fill [yellow,opacity=0.3] (2,4) rectangle (5,9);	
		\fill [yellow,opacity=0.3] (1,5) rectangle (2,8);	
		\fill [yellow,opacity=0.3] (5,5) rectangle (6,8);	
		\fill [yellow,opacity=0.3] (0,6) rectangle (1,7);	
		\fill [yellow,opacity=0.3] (6,6) rectangle (7,7);			
		\draw[step=1.0,gray,thin,opacity=0.5] (0,0) grid (14,10);
		\draw[thick,->] (14,0) -- (0,0) node[above right]{$Y$};
		\draw[thick,->] (14,0) -- (14,10) node[below left]{$X$};
		\draw[black,dashed] (10.5,3.5) circle (2.6);
		\draw[black,dashed] ({3.5+1.3*cos(66.8)},{6.5+1.3*sin(66.8)}) -- ({10.5+1.3*cos(66.8)},{3.5+1.3*sin(66.8)}) -- ({10.5-1.3*cos(66.8)},{3.5-1.3*sin(66.8)}) -- ({3.5-1.3*cos(66.8)},{6.5-1.3*sin(66.8)}) -- cycle;
		\draw[black, thick]
		({3.5+1.3*cos(66.8)+2.252*cos(23.2)},{6.5+1.3*sin(66.8)-2.252*sin(23.2)}) -- ({10.5+1.3*cos(66.8)-2.252*cos(23.2)},{3.5+1.3*sin(66.8)+2.252*sin(23.2)}) -- ({10.5-1.3*cos(66.8)-2.252*cos(23.2)},{3.5-1.3*sin(66.8)+2.252*sin(23.2)}) -- ({3.5-1.3*cos(66.8)+2.252*cos(23.2)},{6.5-1.3*sin(66.8)-2.252*sin(23.2)}) -- cycle;
		\draw[red, thick] (3.5,6.5) circle (2.6);
		\filldraw[black!60!green,->,thick] (6.5,5.5) -- (6.5,7.5);
		\filldraw[black!60!green,->,dashed] (6.5,7.5) -- (5.5,4.5);				
		\filldraw[black!60!green,->,thick] (5.5,4.5) -- (5.5,8.5);
		\filldraw[black!60!green,->,dashed] (5.5,8.5) -- (4.5,3.5);
		\filldraw[black!60!green,->,thick] (4.5,3.5) -- (4.5,9.5);
		\filldraw[black!60!green,->,dashed] (4.5,9.5) -- (3.5,3.5);		
		\filldraw[black!60!green,->,thick] (3.5,3.5) -- (3.5,9.5);
		\filldraw[black!60!green,->,dashed] (3.5,9.5) -- (2.5,3.5);		
		\filldraw[black!60!green,->,thick] (2.5,3.5) -- (2.5,9.5);
		\filldraw[black!60!green,->,dashed] (2.5,9.5) -- (1.5,4.5);		
		\filldraw[black!60!green,->,thick] (1.5,4.5) -- (1.5,8.5);
		\filldraw[black!60!green,->,dashed] (1.5,8.5) -- (0.5,5.5);		
		\filldraw[black!60!green,->,thick] (0.5,5.5) -- (0.5,7.5);
		\filldraw[black!60!green] (6.5,5.5) circle (0.125);
		\draw[black,dashed] (3.5,6.5) -- (10.5,3.5);
		\draw[black,<->] ({3.5+0.3*cos(66.8)},{6.5+0.3*sin(66.8)}) -- ({10.5+0.3*cos(66.8)},{3.5+0.3*sin(66.8)});
		\draw[black] (10.5,3.9) node[above] {$d_{xn}$};
		\draw[black,<->] ({10.5+1.3*cos(66.8)+0.3*cos(23.2)},{3.5+1.3*sin(66.8)-0.3*sin(23.2)}) -- ({10.5-1.3*cos(66.8)+0.3*cos(23.2)},{3.5-1.3*sin(66.8)-0.3*sin(23.2)});
		\draw[black] (10.9,4) node[anchor=north west] {$w_{robot}$};
		\draw[black,<->]  ({3.5+1.3*cos(66.8)+2.252*cos(23.2)+0.2*cos(67.8)},{6.5+1.3*sin(66.8)-2.252*sin(23.2)+0.2*sin(67.8)}) -- ({10.5+1.3*cos(66.8)-2.252*cos(23.2)+0.2*cos(67.8)},{3.5+1.3*sin(66.8)+2.252*sin(23.2)+0.2*sin(67.8)});
		\draw[black] (7,6.5) node[anchor=south west] {$d_{rem}$};
		\draw[black,<->] ({10.5+1.3*cos(66.8)-2.252*cos(23.2)+0.2*cos(67.8)},{3.5+1.3*sin(66.8)+2.252*sin(23.2)+0.2*sin(67.8)}) -- ({10.5+1.3*cos(66.8)+0.2*cos(67.8)},{3.5+1.3*sin(66.8)+0.2*sin(67.8)});
		\draw[black] (10.1,5.0) node[anchor=south west] {$d_{diff}$};
		\filldraw[red] (3.5,6.5) circle (0.25) node[below,black]{\large $\vec{x}_{rand}$};
		\filldraw[blue] (10.5,3.5) circle (0.25) node[anchor=north west,black]{\large $\vec{x}_{n}$};
		\end{tikzpicture}
		\caption{}
		\label{fig:collision2}
	\end{subfigure}
	\caption{Occupancy checks for the grid tiles that intersect with the corridor in (a) and the circle in (b) (red border)  when $\vec{x}_{rand}$ should be connected to $\vec{x}_{n}$. Orange tiles mark the outline found by the proposed methods, while yellow tiles are also checked. The green arrows show the direction of iterating over the tiles. The green rectangle and circle mark the start.}
	\label{fig:collision}
		\vspace{-5mm}
\end{figure}

The set of Nodes $N$ and Edges $E$ will be initialized with the method \texttt{initGraph} which creates a root node at $\vec{x}_{pos}$ and sets $E \leftarrow \emptyset$. The remaining algorithm is executed while the \texttt{explorationFinished()} method, which will be explained in section \ref{exitcondition}, is not satisfied. First, the currently available map $V_{av}$ is retrieved. Then, random samples $\vec{x}_{rand}\in V_{av}$ are generated and if $\vec{x}_{rand}\in V_{free}$, the node closest to each particular sample $\vec{x}_{n}\in N$ is determined.

If the Euclidean distance \texttt{d(}$\vec{x}_{rand},\vec{x}_{n}$\texttt{)}$\ge d_{min}$, $\vec{x}_{n}$ is aligned to the discrete 2D grid map which is required for the \texttt{steer} function. If \texttt{d(}$\vec{x}_{rand},\vec{x}_{n}$\texttt{)}$>d_{max}$, $\vec{x}_{rand}$ is replaced at distance $d_{max}$ to $\vec{x}_{n}$.

Afterwards, all nodes $N_{d}\in N$ in a circle with radius $d_{max}$ around $\vec{x}_{rand}$ are identified. Every node that can be connected to $G$ with the \texttt{steer} function, which will be described in section \ref{steering}, is added to the set $N_{c}$. If $N_{c}\ne\emptyset $, a node at $\vec{x}_{rand}$ is added to the graph with edges to all nodes in $N_{c}$. Otherwise, $\vec{x}_{rand}$ is discarded.

The RRG is built continuously while its nodes are visited by the robot. Compared to rebuilding the graph after reaching a goal, the continuous method can increase total map coverage and prevent getting stuck in a local dead-end where the exploration terminates prematurely.

Furthermore, the random sampling can be extended by sampling in a circular area with radius $ r_{ls} $ around the robot as well. This enhances the local expansion of the tree in large environments which reduces the traveled distance. The local sampling (LS) can be executed in addition to the sampling in all of $V_{av}$, so that two nodes can be added each iteration.

\subsection{Steer Function}\label{steering}

The \texttt{steer} function checks if there is a connection between $\vec{x}_{rand}$ and $\vec{x}_{n}$ that the robot is able to traverse. Since the approach is developed for UGVs, a grid map derived from $V_{av}$ is used to determine traversability. The implemented \texttt{steer} function assumes a non-holonomic robot which is able to turn on the spot.

It requires a circular area around $\vec{x}_{rand}$, which is at least as large as the robot footprint's diagonal $r_{robot}$, and a rectangular corridor between $\vec{x}_{rand}$ and $\vec{x}_{n}$ with at least the robot's width $w_{robot}$, to be traversable. These shapes have to be translated to the grid map's discrete coordinates.

Therefore, the outlines of both shapes are converted to slices in the grid map that are oriented along the $X$ axis. The outlines are derived from the gradients for the corridor and the general form of the equation of a circle. Starting with the rectangular corridor, the algorithm iterates over each grid map tile inside it, from the minimum $x$ value of the particular slice to its maximum $x$ value and from the slice with the smallest $y$ value to the largest $y$ value. For every tile, its occupancy is checked. If a tile is occupied or unknown, \texttt{steer} fails. Otherwise, the same is repeated for the circle

Fig. \ref{fig:collision} shows the process described above. The corridor is shortened to a length $d_{rem}$, so that its width fully intersects the circles to reduce overlapping areas. Otherwise, these areas would be checked twice. $d_{rem}$ is calculated from Equation \eqref{eq:drem}. The circle around $\vec{x}_{n}$ was checked when it was added to the graph and is therefore not required to be checked again.

\begin{equation}
\begin{aligned}
d_{diff}&=\sqrt{r_{robot}^{2}-(\frac{w_{robot}}{2})^{2}} \\
d_{rem}&=\text{d(}\vec{x}_{rand},\vec{x}_{n}\text{)}-2\cdot d_{diff}
\label{eq:drem}
\end{aligned}
\end{equation}

The computational complexity of the \texttt{steer} function is $ \mathcal{O}(g) $ which is derived from the number of cells $ g $ to check in the grid map.

\subsection{Gain function}\label{raycasting}

The Gain $G(n)$ for node $n$ is determined from the expected, additional map coverage at a node's position $\vec{x}_{n}$. This is approximated by the number of voxels in $V_{un}$ that are expected to be sensed at $\vec{x}_{n}$. To carry out this approximation, the sensor's FoV, minimum $r_{min}$ and maximum range $r_{max}$ as well as the sensor's height $h_{sensor}$ above ground must be provided.

To calculate the number of voxels, a method called Sparse Ray Polling (SRP) will be deployed. It is based on Sparse Ray Casting (SRC) proposed by \cite{Selin2019}, which is less computationally intensive than ray tracing in the OctoMap.

For SRP, a set of poll points $P$ is created at the start of the exploration. $P$ is shown in Equation \eqref{eq:pollpoints} where $r_{i}$, $\vartheta_{j}$ and $\varphi_{k}$ are the iterators over the sensor's range, its vertical minimum $\vartheta_{min}$ and maximum FoV $\vartheta_{max}$ as well as a full horizontal revolution from $0$ to $2\pi$. The computational complexity to calculate a node's gain is therefore equal to the number of poll points $|P|=ijk$ to iterate over, resulting in $ \mathcal{O}(ijk)$.

\begin{equation}\label{eq:pollpoints}	
\begin{split}
P & \coloneqq
\bigcup_{r_{i}}
\bigcup_{\vartheta_{j}}
\bigcup_{\varphi_{k}}
\vec{x}_{p}(r_{i},\vartheta_{j},\varphi_{k}) \\
\{r_{i} & =\Delta_{r} \cdot i \, | \, i \in \mathbb{N}, \dfrac{r_{min}}{\Delta_{r}}\leq i\le \dfrac{r_{max}}{\Delta_{r}}\} \\
\{\vartheta_{j} & = \Delta_{\vartheta} \cdot j \, | \, j \in \mathbb{N}, \dfrac{\vartheta_{min}}{\Delta_{\vartheta}}\leq j\le  \dfrac{\vartheta_{max}}{\Delta_{\vartheta}}\} \\
\{\varphi_{k} & = \Delta_{\varphi} \cdot k \, | \, k \in \mathbb{N}, \, 0\leq k < \dfrac{2\pi}{\Delta_{\varphi}}\}		
\end{split}
\end{equation}

The iteration is bound by step sizes $\Delta_{r},\Delta_{\vartheta}$ and $\Delta_{\varphi}$. Too large step sizes cause skipped voxels and wall penetration effects where occupied voxels are omitted and the remaining voxels of the ray are added to the gain. Normally, when an occupied voxel is polled, the remaining poll points on this ray are skipped. If the step sizes are too small, voxels will be polled repeatedly, which increases the computation time and distorts the calculated gain.

To obtain the particular $G(n)$, $P$ is translated to $\vec{x}_{n}$. The translation on the $Z$ axis is assessed by setting the initial node's height $z_{n_i}$ to the average height of all neighbor nodes.

Afterwards, vertical ray tracing in the OctoMap is performed to find the ground's height $h_{ground}$ at $\vec{x}_{n}$. The node's height is then set to $z_n=h_{ground}+h_{sensor}$. If no ground within a maximum height difference $h_{max}$ from $z_{n_i}$ was found, the gain function finishes and $G(n)$ is set to $-1$.

The full horizontal revolution is polled to obtain the best orientation $\vartheta_{max}$ for the robot at $\vec{x}_{n}$ by determining a horizontal slice with the size of the sensor's horizontal FoV with the most gain $G(n)=G(\vartheta_{max})$.

The node's next state $s_{n\texttt{+}1}$ is then derived from Equation \eqref{eq:gain}, which depends on the maximum number of observable voxels in the sensor's FoV $G_{max}$ and a user defined threshold $G_{min}$. If the node's current status $s_n=\textit{visited}$ and the recalculated $\vartheta_{max\texttt{+}1} \approx \vartheta_{max}$, it indicates differences between the approximation and real perception which could lead to the robot getting stuck re-exploring the same node repeatedly.

\begin{equation}
s_{n\texttt{+}1}=\begin{cases}
 \multirow{2}{*}{\textit{explored},} & \text{for } G(n)/G_{max}<G_{min}  \text{ or }\\
 & s_{n}=\textit{visited} \land \vartheta_{max\texttt{+}1} \approx \vartheta_{max}\\
\textit{initial}, & \text{for } G(n)/G_{max}\ge G_{min}\\
\end{cases}
\label{eq:gain}
\end{equation}

\subsection{Cost Function and Path Planning}\label{path}

Cost $C(n)$ for node $n$ is determined by the distance $d_{xn}$ between the node closest to the robot $\vec{n}_{x_{pos}}$ and the particular node's position $\vec{x}_{n}$ along the graph's edges. This distance metric guarantees that a path with the calculated length exists, while the simple Euclidean distance disregards possible obstacles in the way.

The used formula $C(n)\coloneqq e^{-d_{xn}}$ gives a strong bias towards nearby goals. This should reduce the total exploration duration and traveled path length.

$d_{xn}$ and the path $p_{xn}$, on the graph between $\vec{n}_{x_{pos}}$ and $\vec{x}_{n}$, are stored in the proposed approach. Therefore, the robot's position is actively monitored and it is always checked which node in the RRG is closest to the robot. If this node changes, all $d_{xn}$ and $p_{xn}$ are recalculated using Dijkstra's algorithm \cite{Dijkstra1959} starting at  $\vec{n}_{x_{pos}}$. This approach's implementation uses a self-balancing binary search tree which results in a computational complexity of $ \mathcal{O}(|E|\log(|N|)) $.

When a new node $n_{new}$ is added to the graph, $d_{xn_{new}}$ and $p_{xn_{new}}$ are derived from $n_{new}$'s neighbor with the shortest $d_{xn}$. The edge and edge length to this neighbor are added to $p_{xn}$ and $d_{xn}$ respectively and assigned to $n_{new}$. Then, Dijkstra's algorithm is started from $n_{new}$ but without resetting all other $d_{xn}$ and $p_{xn}$ first. Only nodes, which $d_{xn}$ is larger than that of the newly established connection, will therefore be improved. 

\subsection{Exit Conditions}\label{exitcondition}

If the list of nodes ordered by GCR is empty and there is no current goal to pursue, a user defined timer $t_{exit}$ will start. As soon as a new node, which status is not \texttt{explored}, is added to the RRG, the timer will be stopped. If this timer runs out, the exploration is terminated. For an environment with narrow areas, $t_{exit}$ should be increased because random samples are seldom placed in $V_{free}$. The computing power is also a factor to take into consideration, since it influences how long it takes to check if new nodes can be placed.

Furthermore, $G_{min}$ indirectly influences the exit condition by determining, if nodes are set as \texttt{explored}. If $G_{min}$ is set too low, most nodes will serve as a goal for the robot which increases the exploration time. A large value for $G_{min}$ may cause left-out areas in the map on the other hand. 

\section{Implementation}\label{software}

The proposed approach is implemented using ROS and has an interface to the Robot Statemachine (RSM) package \cite{Steinbrink2020} which facilitates its usage.

\begin{figure}[htbp]
	\centerline{\includegraphics[width=0.45\textwidth]{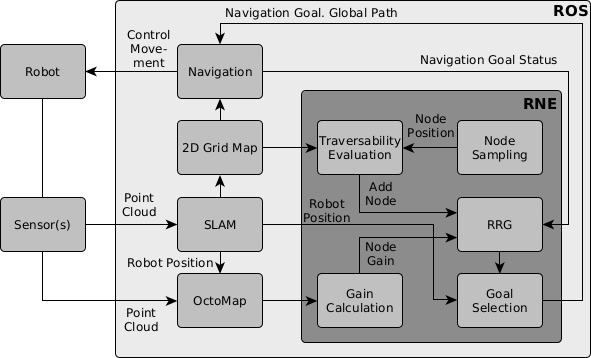}}
	\caption{Overview of RNE's functions interfacing ROS as well as a robot and its sensor(s). RSM is omitted for brevity.}
	\label{fig:rneflow}
	\vspace{-5mm}
\end{figure}

Fig. \ref{fig:rneflow} shows the interaction between RNE and further packages deployed in the robot's ROS environment. Simultaneous Localization and Mapping (SLAM) is used to determine the robot's location and create a 2D grid map which holds traversability information for the ROS navigation package \cite{Marder-Eppstein2010}. The latter receives goals with paths from RNE to control the robot to move along the particular path towards a goal. The robot's position combined with a point cloud from the sensor(s) is used to build an OctoMap \cite{Hornung2013} which is a memory-efficient voxel grid representation.

It is assumed that the robot has a front-facing sensor or a combination of sensors that produces one point cloud within a known range and field of view (FoV). The utilized SLAM algorithm must be able to use the sensor output.

\begin{figure}[htbp]
	\centerline{\includegraphics[width=0.45\textwidth]{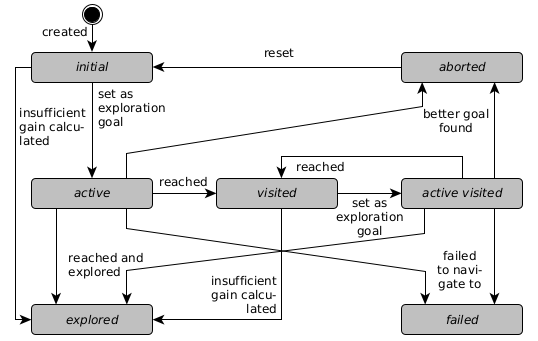}}
	\caption{State diagram of a node's states and their transitions}
	\label{fig:nodestates}
	\vspace{-5mm}
\end{figure}

All nodes in the RRG are assigned one of the seven different states shown in Fig. \ref{fig:nodestates}. When a node is added to the RRG, its gain and cost will be determined by the methods explained in section \ref{raycasting} and \ref{path} respectively and combined using $GCR(n)=G(n)\cdot C(n)$. The nearest nodes and radius searches in the graph are realized with a k-d tree utilizing the efficient nanoflann header-only library \cite{Blanco2014} to interface the RRG.

To obtain the NBV, this approach stores a list of node references ordered by their GCR. This list only contains nodes that are not \textit{explored} or \textit{failed} and which gain has been calculated already. The first node with the best GCR will be the NBV sent to navigation as a goal. When a goal was reached, reaching it failed or another node replaced the current goal node because it has a better GCR, the current goal node's and all nodes' gains inside twice the sensor's range around the robot are recalculated. This only occurs when the robot actually moved towards a goal. Otherwise, it is assumed that the OctoMap did not change and the gains remained the same.

The gain is calculated using computationally intensive queries in the OctoMap and is therefore decoupled from the RRG construction by using a separate thread. All node's, which gains must be calculated, are stored in a list in ascending order by the particular node's distance to the robot, so that the gains closest to the robot are calculated first.

When choosing the next goal, the ordered list of nodes with the best GCR can therefore be incomplete. To enable the robot to start towards a new goal instantaneously, an NBV is selected from the incomplete list nonetheless. As soon as there is another node with a greater GCR, the current goal is aborted and the superior one will be pursued.

\section{Simulations and Experiments}\label{simulations}

RNE introduces additional local sampling (LS) and an RRG implementation with decoupled gain calculation compared to most state-of-the-art RRT implementations. To evaluate if these additions are advantageous, multiple simulations were conducted. Furthermore, our approach as well as adapted versions of RH-NBVP and AEP were deployed in simulated environments for a direct comparison.

\begin{figure*}[!ht]
	\centering
	\begin{subfigure}[t]{.38\textwidth}
		\centering
		\includegraphics[height=3.5cm]{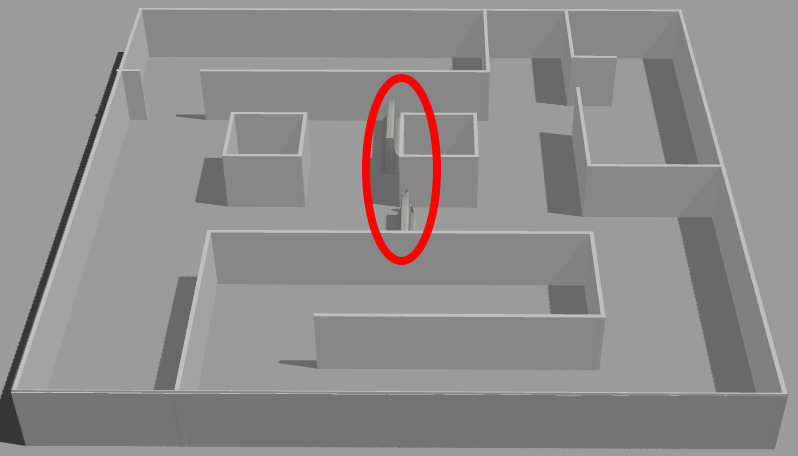}
		\caption{Indoor simulation environment (25x25x2.5m) with omitted roof and added barriers (red circles).}
		\label{fig:gazebo_indoor}
	\end{subfigure}%
	\hfill
	\begin{subfigure}[t]{.38\textwidth}
		\centering
		\includegraphics[height=3.5cm]{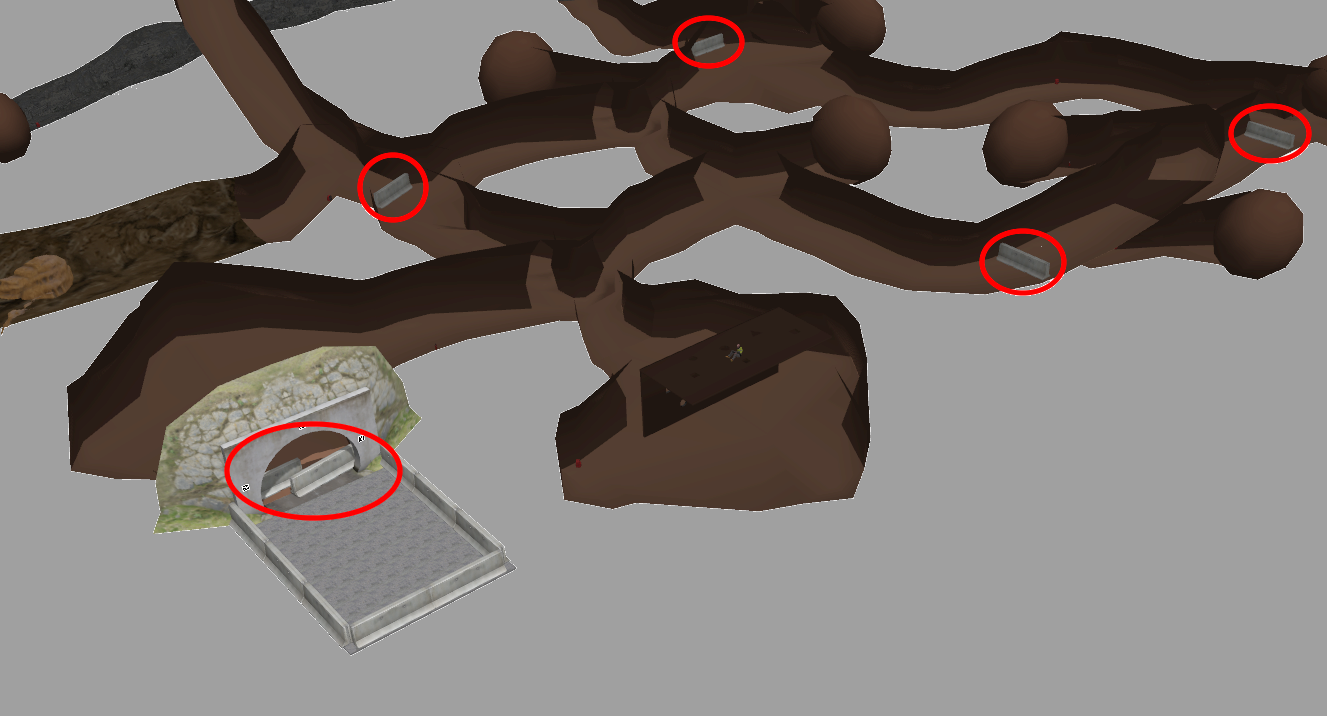}
		\caption{Cave simulation environment (60x90x30m) with added barriers (red circles)}
		\label{fig:gazebo_cave}
	\end{subfigure}%
	\hfill
	\begin{subfigure}[t]{.2\textwidth}
		\centering
		\includegraphics[height=3.5cm]{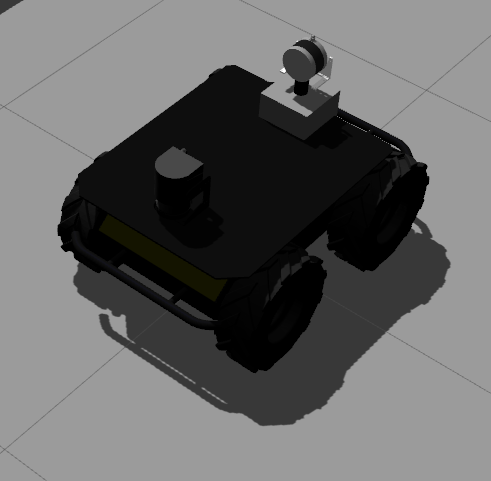}
		\caption{Husky UGV with an externally rotated lidar}
		\label{fig:gazebo_fd}
	\end{subfigure}
	\caption{Gazebo simulation environments shown in (a) and (b) and a robot configuration used in the experiments in (c).}
	\label{fig:gazebo}
	\vspace{-5mm}
\end{figure*}

A computer with Ubuntu 18, a hexa-core 3.2GHz processor and 16GB of RAM was used to run the simulations in Gazebo \cite{Koenig2004}.
Two different robot configurations and three environments were used for the evaluation.

The robot is a Clearpath Robotics Husky UGV equipped with a horizontal lidar in the front, an elevated depth camera in the back for the first configuration (C) and a vertically mounted Velodyne VLP-16 lidar which rotates around the $Z$ axis for the second configuration (L). It is shown in Fig. \ref{fig:gazebo_fd}. The camera's FoV is 87x58 degrees with a range of 8m and the rotated lidar's FoV is 360x135 degrees with a range of 100m.

Fig. \ref{fig:gazebo_indoor} depicts the small (SE) and medium (ME) environments where SE is a small part on the left of ME that is separated by barriers which are removed in the ME scenario. Fig. \ref{fig:gazebo_cave} shows a large underground cave (CE) environment that was modified from \cite{Koval2020} to make untraversable areas inaccessible to the robot. All runs in SE and ME are limited to 30min and runs in CE to 1h.

All simulations, in which the proposed approach is evaluated, use the following parameters: $ t_{exit}=10s, d_{min}=1m, d_{max}=2m, r_{ls}=5m, \Delta_{\varphi}=\Delta_{\vartheta}=10^{\circ}, \Delta_{r}=0.1m, G_{min}=0.05$ for L in SE and ME and $ G_{min}=0.1 $ for every other combination. The ROS package GMapping is used for SLAM. The OctoMap resolution is $e_{V}=0.1m$.

\subsection{Comparison of RRT, RRG and LS}

To show the increased efficiency of our approach, RRG and RRT with and without LS are compared to each other. Four different combinations were executed in 5 variants with 10 runs each. The combinations are RNE (which is RRG with LS), RRG, RRT+LS and RRT. Here, RRT is exactly like our proposed approach, but new nodes are only connected to the nearest node and therefore create a tree. Also, they must be placed at a fixed distance of $d_{min}$.

The variants are C-SE, C-ME, L-SE, L-ME and L-CE. Runs in which the robot gets stuck during navigation are discarded. This is caused by insufficient localization due to erroneous odometry which leads to navigation planning too close to obstacles. A minimum of 7 valid runs is required, otherwise failed runs are repeated. Improving localization to avoid failed runs is out of the scope of this work.

The results of the comparison can be seen in Table \ref{tab:tgls} which shows that RRG is superior to RRT in every scenario regarding the duration and traveled path length. The mean duration of RNE compared to RRT+LS is decreased by up to 48.1\% and the mean path length by up to 43.5\% for L-ME. The mapped volume is approximately equal throughout the runs, but with a decrease for runs that ended prematurely because of the time limit. This causes the standard deviation of 0 for L-CE RRT+LS since all runs stopped because of it.

RRT also has a higher standard deviation for the duration and path length that is caused by the random tree structure in which the robot has to backtrack to reach different branches, while the RRG's interconnected graph leads to more reliable and reproducible explorations.

LS improves the duration and path length for RNE, it has the most significant impact on L-CE. The advantage of RNE is more prominent in larger environments compared to e.g. C-SE where mean duration and distance were only decreased by up to 3\% and 0.1\% respectively.

\begin{figure*}[!ht]
	\centering
	\begin{subfigure}[t]{.24\textwidth}
		\centering
		\includegraphics[width=\textwidth]{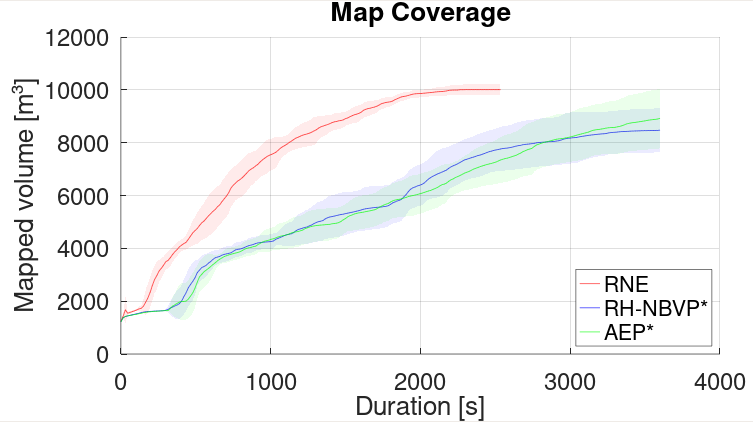}
		\caption{}
	\end{subfigure}
	\hfill
	\begin{subfigure}[t]{.24\textwidth}
		\centering
		\includegraphics[width=\textwidth]{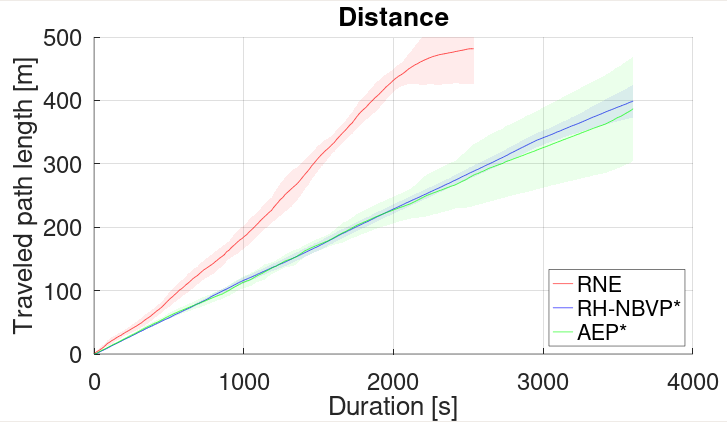}
		\caption{}
	\end{subfigure}
	\hfill
	\begin{subfigure}[t]{.24\textwidth}
		\centering
		\includegraphics[width=\textwidth]{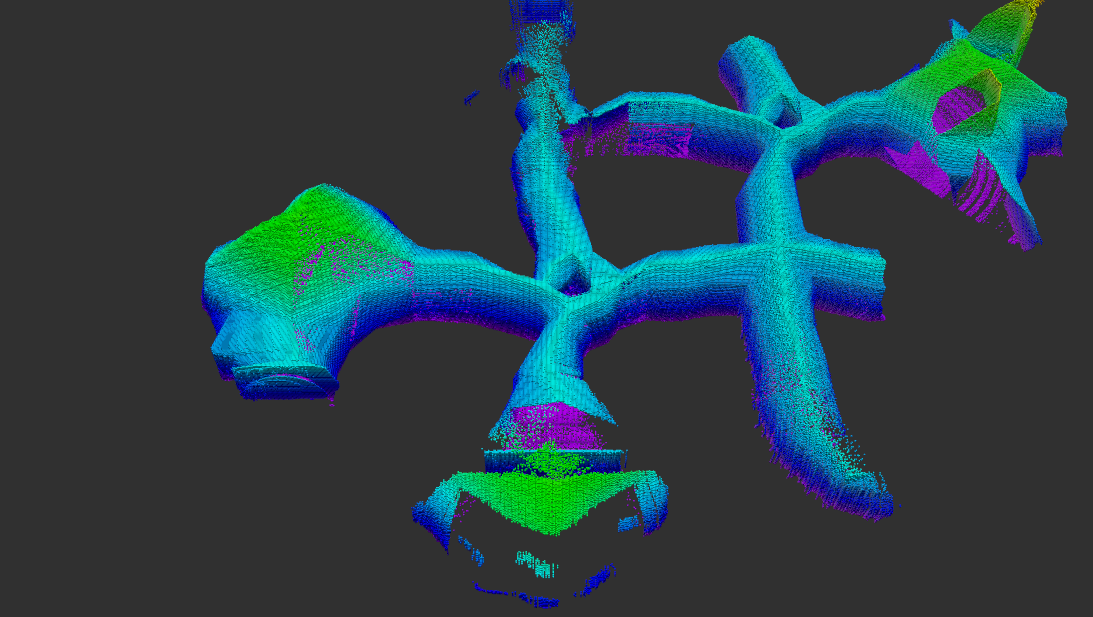}
		\caption{}
	\end{subfigure}
	\hfill
	\begin{subfigure}[t]{.24\textwidth}
		\centering
		\includegraphics[width=\textwidth]{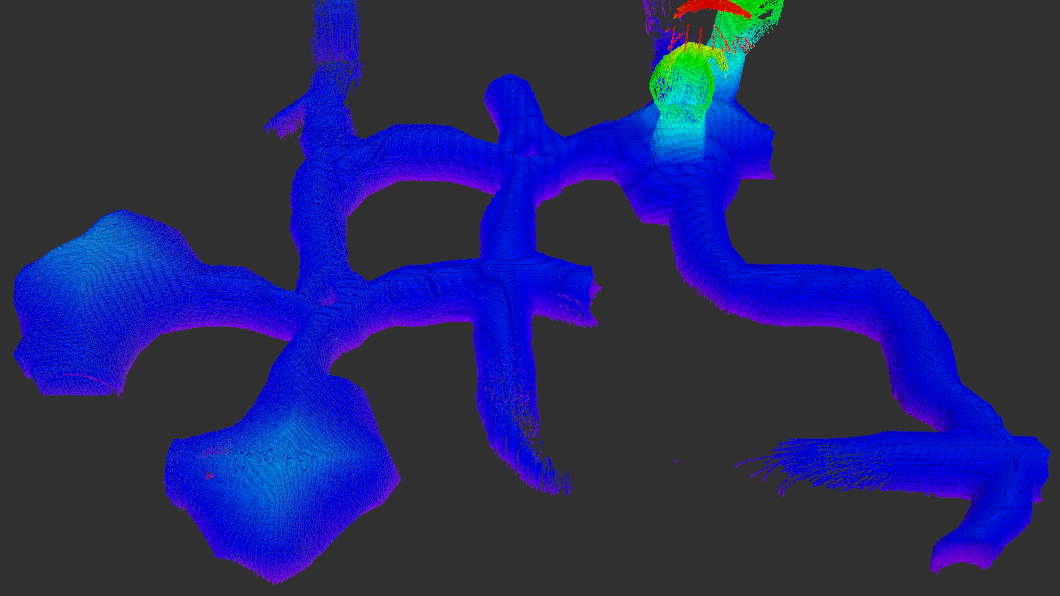}
		\caption{}
	\end{subfigure}
	\caption{Mean mapped volume (a) and path length (b) over time for the L-CE variant. The tinted areas show the standard deviation of the particular values. A line ends at the final duration of the longest run of the particular variant. The recorded OctoMap after 30 minutes of exploring L-CE for AEP* (c) and RNE (d).}
	\label{fig:sota}
	\vspace{-5mm}
\end{figure*}

\subsection{Performance Comparison}

To compare the proposed approach's performance to current state-of-the-art, sampling-based approaches RH-NBVP and AEP, they had to be adapted to use them with the robot configurations presented before. Since they were originally implemented for UAVs, the adaptions use our approach's \texttt{steer} function and 2D sampling method (without LS).

Furthermore, their existing gain functions are replaced with SRP and integrate our exit conditions $ G_{min} $ as well as $ t_{exit} $. For RH-NBVP and AEP, $ t_{exit} $ replaces the maximum tries to find new samples. If the timer runs out, the current best node or frontier for AEP is designated as a goal. If there is no node with a minimum gain, the exploration terminates.  

\begin{table}[htbp]
	\caption{Comparison between RNE, RRG, RRT+LS and RRT showing the mean $\mu$ and standard deviation $\sigma$ of duration, traveled path length and mapped volume.}
	\label{tab:tgls}
	\centering
	\resizebox{\columnwidth}{!}{%
		\begin{tabular}{|c|r|r|r|r|r|r|}
			\hline
			\multirow{2}{*}{\parbox{0.8cm}{Config- uration}} & \multicolumn{2}{l|}{Duration [s]} & \multicolumn{2}{l|}{Path length [m]} & \multicolumn{2}{l|}{Mapped Volume [m\scaleto{\textsuperscript{3}}{5pt}]} \\ \cline{2-7} 
			& $\mu$ & $\sigma$ & $\mu$ & $\sigma$ & $\mu$ & $\sigma$ \\ \thickhline
			C-SE RNE & \textbf{532.50} & 87.73 & \textbf{80.14} & 11.79 & \textbf{803.6} & 14.1 \\ \hline
			C-SE RRG & 549.00 & 58.40 & 80.22 & 8.51 & 799.8 & 15.6 \\ \hline
			C-SE RRT+LS & 798.33 & 157.28 & 113.40 & 30.96 & 768.4 & 87.8 \\ \hline
			C-SE RRT & 836.67 & 323.44 & 129.17 & 32.71 & 798.5 & 23.1 \\ \thickhline
			C-ME RNE & \textbf{1056.67} & 66.29 & \textbf{192.09} & 9.71 & 1685.9 & 7.0 \\ \hline
			C-ME RRG & 1116.67 & 42.50 & 202.31 & 17.83 & \textbf{1692.6} & 9.8 \\ \hline
			C-ME RRT+LS & 1651.88 & 180.26 & 248.18 & 71.44 & 1551.8 & 256.0 \\ \hline
			C-ME RRT & 1713.00 & 164.59 & 276.18 & 32.69 & 1480.8 & 270.6 \\ \thickhline
			L-SE RNE & \textbf{270.00} & 32.40 & \textbf{45.09} & 7.84 & \textbf{928.2} & 23.4 \\ \hline
			L-SE RRG & 319.50 & 57.51 & 57.11 & 8.37 & 913.9 & 24.8 \\ \hline
			L-SE RRT+LS & 370.50 & 76.58 & 72.55 & 10.16 & 910.8 & 26.5 \\ \hline
			L-SE RRT & 372.00 & 60.75 & 70.70 & 9.49 & 917.6 & 32.3 \\ \thickhline
			L-ME RNE & \textbf{768.00} & 93.49 & \textbf{157.86} & 28.09 & 1684.7 & 35.1 \\ \hline
			L-ME RRG & 850.50 & 102.18 & 211.15 & 38.54 & \textbf{1702.5} & 20.0 \\ \hline
			L-ME RRT+LS & 1480.50 & 293.15 & 279.56 & 36.59 & 1701.7 & 57.3 \\ \hline
			L-ME RRT & 1150.50 & 377.51 & 228.57 & 48.44 & 1690.9 & 58.1 \\ \thickhline
			L-CE RNE & \textbf{2155.00} & 199.42 & \textbf{481.41} & 55.97 & 10007.4 & 200.1 \\ \hline
			L-CE RRG & 2655.12 & 217.87 & 682.87 & 72.58 & 10083.9 & 255.5 \\ \hline
			L-CE RRT+LS & 3600.00 & 0.00 & 650.81 & 96.36 & 9946.6 & 505.1 \\ \hline
			L-CE RRT & 3551.67 & 145.00 & 897.58 & 82.35 & \textbf{10296.8} & 44.8 \\ \hline
		\end{tabular}%
	}
\end{table}

Because of these adaptions, the two approaches will be referenced as RH-NBVP* and AEP* in the following. The same five variants as in the previous simulations are executed 10 times each. The RRT's maximum edge length for both is $l=1m$, RH-NBVP*'s degression coefficient is set to $\lambda=0.5$ and AEP*'s to $\lambda=0.75$, its GCR threshold to $g_{zero}=1$. The maximum and minimum amount of nodes are $N_{max}=400, N=30$ for ME and CE and $N_{max}=200, N=15$ for SE for RH-NBVP*. AEP*'s $N$ is the same and $N_{max}$ is only half of RH-NBVP*'s. These values are based on \cite{Selin2019}.

Table \ref{tab:sota} shows the results of RH-NBVP* and AEP*. RNE is listed again for better comparability. Furthermore, Fig. \ref{fig:sota} displays the mean volume and the path length over time for RNE, AEP* and RH-NBVP* in the L-CE variant as well as the OctoMap after 30 minutes of exploration for AEP* and RNE. The continuously-built RRG with LS and the decoupled gain calculation lead to vaster explored areas of the map in less time while traveling shorter distances.

The proposed RNE achieves an increase in mapped volume of 18.1\% compared to RH-NBVP* and 12\% to AEP* in the L-CE variant, while finishing the exploration in 38.8\% and 37.1\% less time respectively. The path lengths of RH-NBVP* and AEP* are shorter because of the time limit, at which they are still not finished with the exploration, and their slower pace compared to RNE. Even in the smaller C-SE variant, it decreased the duration by 14.6\% compared to RH-NBVP* and 19.1\% to AEP* as well as the distance by

\begin{table}[htbp]
	\caption{RNE, RH-NBVP* and AEP* simulation results with mean $\mu$ and standard deviation $\sigma$ of duration, traveled path length and mapped volume.}
	\label{tab:sota}
	\centering
	\resizebox{\columnwidth}{!}{%
		\begin{tabular}{|c|r|r|r|r|r|r|}
			\hline
			\multirow{2}{*}{\parbox{0.8cm}{Config- uration}} & \multicolumn{2}{l|}{Duration [s]} & \multicolumn{2}{l|}{Path length [m]} & \multicolumn{2}{l|}{Mapped Volume [m\scaleto{\textsuperscript{3}}{5pt}]} \\ \cline{2-7} 
			& $\mu$ & $\sigma$ & $\mu$ & $\sigma$ & $\mu$ & $\sigma$ \\ \thickhline
			C-SE RNE & \textbf{532.50} & 87.73 & \textbf{80.14} & 11.79 & \textbf{803.6} & 14.1 \\ \hline
			C-SE RH-NBVP* & 623.33 & 96.66 & 89.33 & 16.18 & 746.0 & 7.0 \\ \hline
			C-SE AEP* & 658.50 & 100.26 & 100.61 & 14.28 & 756.0 & 12.7 \\ \thickhline
			C-ME RNE & \textbf{1056.67} & 66.29 & \textbf{192.09} & 9.71 & \textbf{1685.9} & 7.0 \\ \hline
			C-ME RH-NBVP* & 1782.86 & 45.36 & 250.79 & 63.73 & 1494.3 & 245.8 \\ \hline
			C-ME AEP* & 1762.50 & 59.37 & 256.19 & 17.78 & 1575.0 & 87.9 \\ \thickhline
			L-SE RNE & \textbf{270.00} & 32.40 & \textbf{45.09} & 7.84 & \textbf{928.2} & 23.4 \\ \hline
			L-SE RH-NBVP* & 820.71 & 170.08 & 111.61 & 23.67 & 922.3 & 21.4 \\ \hline
			L-SE AEP* & 589.29 & 64.64 & 86.11 & 14.77 & 911.9 & 25.1 \\ \thickhline
			L-ME RNE & \textbf{768.00} & 93.49 & \textbf{157.86} & 28.09 & \textbf{1684.7} & 35.1 \\ \hline
			L-ME RH-NBVP* & 1786.88 & 24.63 & 234.72 & 5.63 & 1574.3 & 106.9 \\ \hline
			L-ME AEP* & 1386.43 & 346.07 & 180.43 & 74.59 & 1479.5 & 204.9 \\ \thickhline
			L-CE RNE & \textbf{2155.00} & 199.42 & 481.41 & 55.97 & \textbf{10007.4} & 200.1 \\ \hline
			L-CE RH-NBVP* & 3519.38 & 228.04 & 398.83 & 25.78 & 8471.9 & 837.9 \\ \hline
			L-CE AEP* & 3429.00 & 530.29 & \textbf{387.44} & 81.84 & 8934.4 & 1137.7 \\ \hline
		\end{tabular}%
	}
\end{table}

\noindent 10.3\% and 20.3\% respectively.

A video of RNE on a robot with the rotating Velodyne lidar exploring the simulated cave environment is provided here: \url{https://youtu.be/00UQ3JTSeX8}.

\section{Conclusion}

In this paper, we presented an RRG-based approach that outperforms state-of-the-art, sampling-based methods in the following aspects. First, our simulations show the advantages of RRG compared to RRT regarding run duration and traveled path length. Second, LS additionally decreased the duration and path length of the exploration. Finally, we demonstrated the superior efficiency of our novel approach, which employs a persistent RRG with LS and decoupled gain calculation, compared to RH-NBVP* and AEP*. 

Future research will focus on optimizing the potential gain evaluation to more precisely decide if a node is worth exploring. Therefore, ray clustering for SRP will be researched. Also, a 6D SLAM and traversability evaluation will be deployed to explore more difficult environments.
 

\bibliography{./library}
\bibliographystyle{IEEEtran}

\end{document}